\pdfoutput=1
\documentclass[11pt]{article}
\usepackage[preprint]{acl}
\usepackage{times}
\usepackage{latexsym}
\usepackage[T1]{fontenc}
\usepackage[utf8]{inputenc}
\usepackage{microtype}
\usepackage{graphicx}
\usepackage{booktabs}
\usepackage{amsmath}
\usepackage{xurl}
\usepackage{float}
\usepackage{placeins}
\usepackage{stfloats}
\usepackage{array}
\usepackage{enumitem}
\usepackage{listings}
\usepackage{needspace}
\newcommand{\FormWithoutPrev}{8.7}
\newcommand{\FormSharedPrev}{42}
\newcommand{\FormWithoutBase}{8.4}
\newcommand{\FormSharedBase}{50}
\newcommand{\FormWithoutNext}{8.2}
\newcommand{\FieldRemovedChange}{-1.1}
\newcommand{\QuestionsRemovedChange}{-0.2}
\newcommand{\FlagGapEarly}{3.0}
\newcommand{\FeGapEarly}{2.7}
\newcommand{\ControlGapEarly}{3.3}
\newcommand{\FlagGapLate}{4.5}
\newcommand{\FeGapLate}{7.7}
\newcommand{\ControlGapLate}{7.2}
\newcommand{\PlaceboFirstGap}{-3.3}
\newcommand{\PlaceboMidGap}{-1.2}
\newcommand{\PlaceboLastGap}{0.2}
\newcommand{\JournalGap}{5.4}
\newcommand{\JournalGapVarying}{2,228}
\newcommand{\MeldThreshold}{2.75}
\newcommand{\MeldCalibrationReviews}{110,080}
\newcommand{\MeldIntelAiReviews}{5,000}
\newcommand{\MeldPreSourceMax}{0.4}
\newcommand{\MeldIntelAiRecall}{100.0}
\newcommand{\IcmlFlaggedBase}{35.5}
\newcommand{\CrossRecentLast}{0.2}
\newcommand{\CrossLast}{0.1}
\newcommand{\EmbedScaledSameBase}{0.53}
\newcommand{\EmbedScaledSameMid}{0.57}
\newcommand{\EmbedScaledSamePrev}{0.65}
\newcommand{\EmbedScaledSameLast}{0.66}
\newcommand{\ClarityBase}{24}
\newcommand{\ClarityLast}{10}
\newcommand{\TyposBase}{6}
\newcommand{\TyposLast}{1}
\newcommand{\ExperimentsBase}{13}
\newcommand{\ExperimentsLast}{18}
\newcommand{\RigorBase}{2}
\newcommand{\RigorLast}{7}
\newcommand{\FResearchChange}{5.8}
\newcommand{\PlosChange}{2.5}
\newcommand{\RefPanels}{510}
\newcommand{\RefCrossPanels}{90}
\newcommand{\RefReviews}{2106}
\newcommand{\RefSpanFOne}{78.5}
\newcommand{\RefCouncilSpanFOne}{80.5}
\newcommand{\RefSpecific}{89.7}
\newcommand{\RefPairPanels}{507}
\newcommand{\RefPairTp}{1,148}
\newcommand{\RefPairFp}{226}
\newcommand{\RefPairFn}{316}
\newcommand{\RefPairCandidates}{141,426}
\newcommand{\RefPairAllCandidates}{141,700}
\newcommand{\RefPairCoverage}{99.8}
\newcommand{\RefPairPrecision}{83.6}
\newcommand{\RefPairRecall}{78.4}
\newcommand{\RefRatioPlain}{87}
\newcommand{\RefRatioFlagged}{101}
\newcommand{\RefTypeRatioNone}{100}
\newcommand{\RefTypeRatioTwo}{101}
\newcommand{\RefTypeLinksNone}{70}
\newcommand{\RefSwapPanels}{405}
\newcommand{\RefSwapFlagLuna}{7.4}
\newcommand{\RefSwapFlagMin}{7.0}
\newcommand{\RefSwapFlagMax}{7.3}
\newcommand{\RefSwapPanelLuna}{5.9}
\newcommand{\RefSwapPanelMin}{3.9}
\newcommand{\RefSwapPanelMax}{5.1}
\newcommand{\RefPipePanels}{510}
\newcommand{\RefPairEmptyPanels}{3}
\newcommand{\RefPipeLuna}{11.1}
\newcommand{\RefPipeAbsDiffLuna}{5.5}
\newcommand{\RefPipeAbsDiffCouncil}{5.5}
\newcommand{\RefPipeLevelMemberMin}{10.9}
\newcommand{\RefPipeLevelMemberMax}{14.8}
\newcommand{\RefPipeStepSignificant}{4}
\newcommand{\RefPipePairGapSignificant}{3}
\newcommand{\KimReviews}{218}
\newcommand{\KimItems}{1,930}
\newcommand{\KimItemRecall}{96.7}
\newcommand{\KimItemSplit}{30.0}
\newcommand{\KimClaimUnmatched}{0.7}
\newcommand{\KimClaimsBias}{3.0}
\newcommand{\ClaimcheckReviews}{60}
\newcommand{\ClaimcheckItems}{168}
\newcommand{\ClaimcheckItemRecall}{94.2}
\newcommand{\IdJournalDirect}{5.0}
\newcommand{\IdJournalDirectN}{230}
\newcommand{\RobustMin}{2.5}
\newcommand{\RobustMax}{4.1}
\newcommand{\IclrSnapBase}{12.0}
\newcommand{\IclrSnapLast}{21.3}
\newcommand{\IclrSnapChange}{9.3}
\newcommand{\IclrKeptChange}{9.6}
\newcommand{\WithdrawnShare}{29}
\newcommand{\WithdrawnEarlyShare}{9}
\newcommand{\WithdrawnPapers}{5,715}
\newcommand{\WithdrawnLevel}{13.1}
\newcommand{\RobustSnapMin}{8.9}
\newcommand{\RobustSnapMax}{10.4}
\newcommand{\SizeLlSmallerEarly}{-0.82}
\newcommand{\DecSnapChange}{9.3}
\newcommand{\DecSnapBackground}{3.7}
\newcommand{\DecSnapLlm}{5.6}
\newcommand{\DecSnapNoneLast}{14.5}
\newcommand{\DecIclrNoneLast}{10.8}
\newcommand{\DecIclrCleanBase}{9.0}
\newcommand{\DecIclrCleanLast}{10.3}
\newcommand{\DecSnapCleanLast}{14.3}
\newcommand{\CovProblemsLast}{19.9}
\newcommand{\CovPerReviewLast}{7.2}
\newcommand{\CovNewLast}{6.12}
\newcommand{\CovProblemsSnap}{19.0}
\newcommand{\CovPerReviewSnap}{7.3}
\newcommand{\CovNewSnap}{5.46}
\newcommand{\RwPapers}{800}
\newcommand{\RwIclrPapers}{600}
\newcommand{\RwIclrTwo}{0.9}
\newcommand{\RwIclrUnchanged}{0.3}
\newcommand{\RwLunaRecall}{86.2}
\newcommand{\PlIclrTwo}{0.5}
\newcommand{\PlLunaRecall}{94.7}
\newcommand{\RkIclrPapers}{600}
\newcommand{\PolishTexts}{2,100}
\newcommand{\PolishFlagged}{39.5}
\newcommand{\PolishUnflagged}{60.5}
\newcommand{\RewriteTexts}{13,194}
\newcommand{\RewriteUnflagged}{0.8}
\newcommand{\GigacheckThreshold}{8.01}
\newcommand{\GigacheckAgreementReviews}{415,248}
\newcommand{\GigacheckKappa}{0.34}
\newcommand{\GigacheckBoth}{12.2}
\newcommand{\GigacheckMeldOnly}{24.9}
\newcommand{\GigacheckOnly}{2.1}
\newcommand{\GigacheckHeldoutFlagged}{0.08}
\newcommand{\GigacheckGapIclrEarly}{2.2}
\newcommand{\GigacheckGapIclrLate}{5.4}
\newcommand{\GigacheckGapNeurips}{5.1}
\newcommand{\GigacheckGapIcml}{5.5}
\newcommand{\ReproCalibrationInAnalysis}{94,706}
\newcommand{\ReproCalibrationOutside}{15,374}

\title{Twice-Told Critiques: Claim-Level Homogenization of Peer Review in the LLM Era}
\author{Ilya Gusev \\ Keenable / Amsterdam \\ \texttt{phoenixilya@gmail.com}}

\newcounter{rq}
\newcommand{\question}[2]{\refstepcounter{rq}\paragraph{RQ\therq: #2}\label{#1}}

\begin{document}
\maketitle

\begin{abstract}
Peer review relies on several readers finding different problems. We measure critique overlap in conference and journal reviews. At ICLR, measured overlap rose from \IclrSnapBase\% in 2026 to \IclrSnapLast\% in 2027, in reviews from the same stage of the review process, with year-on-year rises at NeurIPS, ICML and the F1000 and PLOS journals. A third ICLR review of a paper added \CovNewSnap{} new problems to the other two in 2027, against \CovNewLast{} in 2026. We apply MELD, a detector of LLM-generated text, to all reviews. Within a paper, two reviews that MELD flags overlap more than two unflagged reviews: by \FlagGapEarly{} points at ICLR in 2024--2025, by \FlagGapLate{} points in 2026--2027, and by \JournalGap{} points across five journal sources. These results provide evidence that LLM-flagged reviews of one paper are less distinct from each other than unflagged reviews, and suggest that the diversity of critiques among reviews of a single paper is falling.
\end{abstract}

\begin{figure}[t]
\centering
\includegraphics{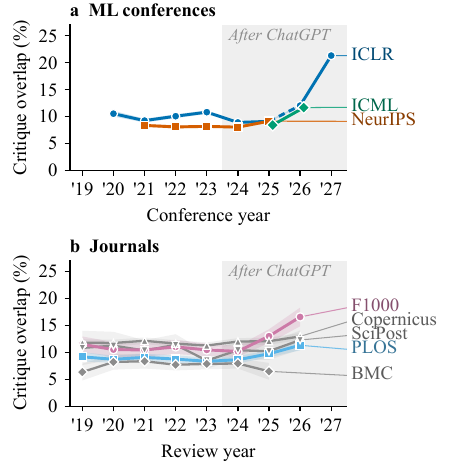}
\caption{Measured critique overlap rose at ICLR in 2027, at ICML in 2026 and at NeurIPS in 2025. Journal overlap rose most at F1000 and PLOS. We define overlap as shared problems divided by the smaller critique set, averaged within papers, then over papers. Dashed segment: from 2026 on, ICLR samples hold reviews during the discussion phase. The 2026 and 2027 points include papers withdrawn later.}
\label{fig:overlap}
\end{figure}

\section{Introduction}

Peer review needs readers who find different problems. In one study, two reviewers usually shared no critical point \citep{fiske1990reviewers}. In another study about Nature-family reviews, 25.8\% of one reviewer's criticisms match another's \citep{kim2026limits}. Our central question is: with widespread LLM-generated text in ML reviews \citep{liang2024monitoring,russo2025lottery,emi2025pangram}, do reviewers still provide distinct critiques?

We measure shared problems by extracting and matching critiques across reviews of one paper. Whole-review similarity mixes wording and content \citep{liang2024monitoring,baumann2026stop} and cannot establish that two reviews raise the same problem. Prior studies find that synthetic LLM-generated reviews are more similar to each other than human reviews \citep{hou2026converge} and share more criticisms \citep{kim2026limits}, a gap used for detection \citep{duarte2026semdetect}. We apply critique extraction \citep{liang2024feedback,kim2026limits} to reviews that venues received, comparing problems within papers and over time.

Our findings are:

\begin{itemize}[nosep]
\item Critique overlap rose at ICLR from \IclrSnapBase\% in 2026 to \IclrSnapLast\% in 2027, at ICML in 2026 and at NeurIPS in 2025 (Section~\ref{sec:trends}, Figure~\ref{fig:overlap}).
\item Within a paper, two LLM-flagged reviews overlap more than two unflagged reviews at ICLR, by \FlagGapEarly{} points in 2024--2025 and \FlagGapLate{} points in 2026--2027 (Section~\ref{sec:ai}, Figure~\ref{fig:ai}).
\item Pooled over five journal sources, the same within-paper gap is \JournalGap{} points (Section~\ref{sec:journals}, Figure~\ref{fig:fields}).\label{end:intro}
\end{itemize}

\section{Related Work}
\label{sec:related}

\paragraph{LLM use in peer review.} ICLR and NeurIPS reviews grew longer after ChatGPT, while LLM-assisted ICLR reviews mention originality less \citep{wu2026impact}. LLM reviewers draw more than half of their weaknesses from the appendix. In sampled ICLR reviews, the share of such weaknesses rose from 11.3\% in 2026 to 27.6\% in 2027 \citep{ghosh2026appendix}. LLM-generated reviews at one management journal cover a narrower range of topics \citep{gartenberg2026more}. At ICML 2026, a randomized trial assigned papers and reviewers to a policy that prohibited LLM use or to one that allowed limited use. The second policy made reviews 5.5--7\% longer and left scores and decisions almost unchanged \citep{kim2026icmltrial}. We compare shared problems, not single-review topics.

\paragraph{Homogenization.} LLM assistance makes the texts and ideas of different users more alike \citep{padmakumar2024writing,anderson2024homogenization,doshi2024generative}. Model outputs are narrower than training data \citep{wu2025monoculture}. \citet{sourati2026homogenizing} review this evidence. We extend the question from creative tasks to expert critiques, where independent readings are the goal.

\section{Methodology}
\label{sec:data}

\subsection{Data}
Table~\ref{tab:data} lists the samples. ICLR and NeurIPS come from OpenReview \citep{arcade2025dataset,xu2025researcharcade,heineman2026iclr}. ICML includes accepted papers from 2025--2026. ICLR 2027 reviews come from the OpenReview API eight days after they became public. Conference papers need at least three reviews. Journal papers need at least two first-round reviews: SciPost, F1000Research and sister platforms (F1000), PLOS, BMC Medicine \citep{jerma2026bmc}, and Copernicus/EGU, covering physics, medicine, biology and geoscience. A \textbf{panel} is the set of reviews of one paper.

\begin{table}[t]
\centering\footnotesize
\begin{tabular}{llrr}
\toprule
Source & Years & Papers & Reviews \\
\midrule
ICLR & 2020--27 & 85,466 & 299,168 \\
NeurIPS & 2021--25 & 18,746 & 74,866 \\
ICML & 2025--26 & 9,588 & 36,804 \\
SciPost & 2019--26 & 2,962 & 6,613 \\
F1000 & 2019--26 & 7,111 & 16,983 \\
PLOS & 2019--26 & 14,360 & 34,084 \\
BMC & 2019--25 & 2,510 & 6,414 \\
Copernicus & 2019--26 & 16,000 & 34,877 \\
\bottomrule
\end{tabular}

\caption{Papers and reviews in the analysis.}
\label{tab:data}
\end{table}

\subsection{Claim extraction}
\label{sec:extraction}
A \textbf{critique claim} identifies one problem that needs one fix. We exclude praise and neutral text. In \textbf{pass 1}, GPT-6 Luna (Luna) reads one review in full and lists its claims. Each claim has a quote copied from the review, a one-sentence paraphrase and a category such as clarity or experiment scope.

We check pass 1 against human labels and against other LLMs. For the human check, we use two public sets: \KimReviews{} Nature-family reviews that the authors of the set split into \KimItems{} critique items \citep{kim2026limits}, and \ClaimcheckReviews{} NeurIPS reviews with \ClaimcheckItems{} weakness spans that experts marked \citep{ou2025claimcheck}. A quote of at least one Luna claim falls in \KimItemRecall\% of the items and in \ClaimcheckItemRecall\% of the spans, and \KimClaimUnmatched\% of Luna's claims on the first set fall outside every item. For the LLM check, Sol 6.1, DeepSeek V4.1 Flash and Kimi K3.1 Pro run pass 1 on \RefReviews{} of our reviews, in \RefPanels{} \textbf{validation panels} from every venue, year and detector stratum. We call these three models the \textbf{council}. Luna's quotes reach a word-span F1 of \RefSpanFOne\% with a council member, against \RefCouncilSpanFOne\% between two members.

\subsection{Claim grouping}
\label{sec:grouping}
In \textbf{pass 2}, a separate Luna call groups the claims of one panel into problems. The call sees the claim lists, not the review texts. Two claims join one group only when they state the same problem in the same paper element.

To check pass 2, each council member groups Luna's claim lists with the same prompt. Against the majority vote of the three members on each claim pair, Luna's grouping has \RefPairPrecision\% precision and \RefPairRecall\% recall across \RefPairPanels{} panels. With a council member or the majority as grouper, the within-paper flagged-pair gap of RQ2 is \RefSwapFlagMin{} to \RefSwapFlagMax{} points, against \RefSwapFlagLuna{} with Luna (Table~\ref{tab:swap}). Each council member also runs both passes on its own claims, and all four pipelines give every contrast the same sign (Table~\ref{tab:pipelines}).

\begin{table}[t]
\centering\footnotesize
\begin{tabular*}{\columnwidth}{@{\extracolsep{\fill}}lrr@{}}
\toprule
 & \multicolumn{2}{c}{Within-paper gap (points)} \\
\cmidrule(l){2-3}
Grouper & One flagged & Two flagged \\
\midrule
GPT-6 Luna & +2.9 [0.2, 5.6] & +7.4 [3.6, 11.3] \\
Sol 6.1 & +2.1 [$-$0.7, 4.9] & +7.1 [3.3, 10.9] \\
DeepSeek V4.1 Flash & +1.8 [$-$1.8, 5.4] & +7.3 [2.7, 11.8] \\
Kimi K3.1 Pro & +2.7 [$-$0.8, 6.1] & +7.2 [2.7, 11.7] \\
Council majority & +2.1 [$-$1.3, 5.5] & +7.0 [2.7, 11.2] \\
\bottomrule
\end{tabular*}

\caption{Every grouper gives about the same within-paper gaps. Each grouper groups Luna's claims on \RefSwapPanels{} validation panels from 2024 on. Gap: overlap of pairs with one or two flagged reviews minus unflagged pairs of the same paper, with 95\% intervals.}
\label{tab:swap}
\end{table}

\begin{table}[t]
\centering\footnotesize
\setlength{\tabcolsep}{3pt}
\begin{tabular*}{\columnwidth}{@{\extracolsep{\fill}}lrrrr@{}}
\toprule
 & \multicolumn{2}{c}{Overlap (\%)} & \multicolumn{2}{c}{Contrast (points)} \\
\cmidrule(lr){2-3}\cmidrule(l){4-5}
Pipeline & \shortstack[r]{Same\\paper} & \shortstack[r]{Cross\\paper} & \shortstack[r]{Flagged\\gap} & \shortstack[r]{ICLR 2027\\$-$ 2026} \\
\midrule
GPT-6 Luna & 11.1 & 0.4 & +7.4 & +7.0 \\
Sol 6.1 & 10.9 & 0.5 & +6.1 & +9.0 \\
DeepSeek V4.1 Flash & 14.8 & 0.5 & +4.2 & +7.9 \\
Kimi K3.1 Pro & 14.7 & 1.0 & +10.0 & +9.6 \\
\bottomrule
\end{tabular*}

\caption{All four pipelines give every contrast the same sign; the overlap level depends on the model. Each model extracts and groups its own claims on \RefPipePanels{} same-paper and \RefCrossPanels{} cross-paper validation panels. Flagged gap: the two-flagged gap of Table~\ref{tab:swap}.}
\label{tab:pipelines}
\end{table}

\subsection{Metric}
\label{sec:metric}
Let $G_A$ and $G_B$ be the sets of distinct problems that reviews $A$ and $B$ raise. Their \textbf{critique overlap} is the overlap coefficient \citep{simpson1960notes}
\begin{equation}
 o(A,B)=\frac{|G_A\cap G_B|}{\min(|G_A|,|G_B|)},
\label{eq:overlap}
\end{equation}
and $o(A,B)=0$ for an empty set. Full overlap means containment of the smaller set, not identical reviews. Higher overlap can result from smaller critique sets rather than more shared problems. Every main claim keeps its sign under other measures, including shared-problem counts at fixed set sizes. Following the IntraSim measure of \citet{baumann2026stop}, we average $o$ over all review pairs within a paper, then over papers within a year.

\subsection{Detector of LLM-generated texts}
\label{sec:detector}
MELD \citep{li2026meld} scores reviews for LLM-generated or edited text. \textbf{LLM-flagged} means above one common threshold: the 99.9th percentile of \MeldCalibrationReviews{} pre-2023 scores. MELD flags at most \MeldPreSourceMax\% of each source's pre-2023 reviews and \MeldIntelAiRecall\% of \MeldIntelAiReviews{} generated ICLR reviews \citep{yu2026reviewed}. MELD flags \PolishFlagged\% of Luna-polished human reviews.

\begin{figure}[t]
\centering
\includegraphics{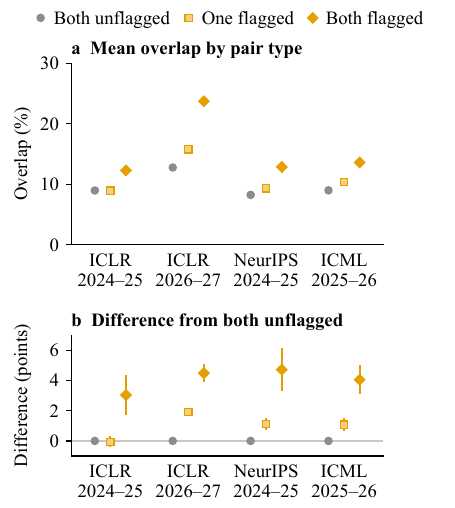}
\caption{In both ICLR periods, at NeurIPS and at ICML, two LLM-flagged reviews of one paper overlap more than two unflagged reviews. A pair type counts the flagged reviews of a pair. (a) Mean over papers of pair overlap within each paper. (b) Mean over papers of the difference from the unflagged pairs of the same paper. Bars: 95\% intervals.}
\label{fig:ai}
\end{figure}

\section{Results}
\label{sec:results}

\paragraph{RQ1: Do reviewers of one paper repeat more of each other's critiques?}
\label{sec:trends}
ICLR overlap rose from \IclrSnapBase\% in 2026 to \IclrSnapLast\% in 2027, by $+\IclrSnapChange$ points. These two years are directly comparable: both hold reviews during the discussion phase. NeurIPS and ICML also rose in their latest cycles (Figure~\ref{fig:overlap}).

Table~\ref{tab:coverage} counts claims in all three-review subsets of each ICLR paper. From 2026 to 2027, claims per review rose from \CovPerReviewLast{} to \CovPerReviewSnap, but distinct claims per three-review panel fell from \CovProblemsLast{} to \CovProblemsSnap, and the new claims that a third review added fell from \CovNewLast{} to \CovNewSnap.

\begin{table}[t]
\centering\footnotesize
\setlength{\tabcolsep}{0pt}
\begin{tabular*}{\columnwidth}{@{\extracolsep{\fill}}lcccc@{}}
\toprule
 & \multicolumn{4}{c}{Claims} \\
\cmidrule{2-5}
Year & \shortstack{Per\\review} & \shortstack{Distinct} & \shortstack{New in a\\third review} & \shortstack{Shared\\per pair} \\
\midrule
2020 & 7.02 & 19.73 & 6.16 & 0.48 \\
2021 & 7.65 & 21.62 & 6.79 & 0.48 \\
2022 & 7.57 & 21.32 & 6.68 & 0.49 \\
2023 & 7.06 & 19.79 & 6.17 & 0.49 \\
2024 & 6.87 & 19.44 & 6.11 & 0.42 \\
2025 & 7.20 & 20.34 & 6.38 & 0.45 \\
2026\rlap{$^\dagger$} & 7.20 & 19.92 & 6.12 & 0.61 \\
2027\rlap{$^\dagger$} & 7.31 & 18.98 & 5.46 & 1.10 \\
\bottomrule
\end{tabular*}

\caption{ICLR three-review panels hold fewer distinct claims in 2027 than in any year from 2020, although claims per review did not fall from 2026. Distinct: claims of the three reviews, with a repeated claim counted once. $^\dagger$Reviews during the discussion phase; earlier years hold final reviews.}
\label{tab:coverage}
\end{table}

\paragraph{RQ2: Do LLM-flagged reviews share a larger fraction of their critiques?}
\label{sec:ai}
For each paper with both pair types, we subtract the mean overlap of its pairs of two unflagged reviews from the mean overlap of its pairs of two flagged reviews. The \textbf{flagged-pair gap} is the mean of this difference over papers. At ICLR, the flagged-pair gap is \FlagGapEarly{} points in 2024--2025 and \FlagGapLate{} points in 2026--2027. We do not pool them: their reviews come from different review stages. The gap is also positive at NeurIPS and at ICML (Figure~\ref{fig:ai}).

If the overlap of every ICLR pair had risen from 2026 to 2027 as much as the overlap of two unflagged reviews, from \DecIclrNoneLast\% to \DecSnapNoneLast\%, the total would have risen by \DecSnapBackground{} points, not \DecSnapChange. The other \DecSnapLlm{} points go with flagged reviews. This split is an accounting, not an effect of LLM use.

The gap could come from wording and not from content, if two reviews in one LLM style are easier to match. To test this style effect, Luna changes two human reviews in each of \RwIclrPapers{} pre-2023 ICLR panels and keeps their points. Polishing the language changes overlap by $+\PlIclrTwo$ points, and rewriting each review from an outline of its points by $+\RwIclrTwo$ points, against ICLR flagged-pair gaps of \FlagGapEarly{} and \FlagGapLate{} points. Style effects can therefore explain only part of the flagged-pair gap.

\begin{figure}[t]
\centering
\includegraphics{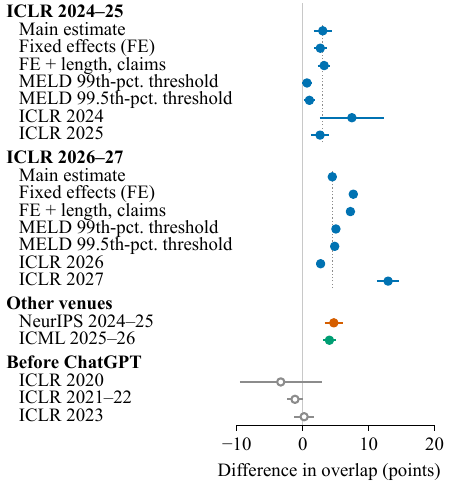}
\caption{The within-paper flagged-pair gap is positive in both ICLR periods, at NeurIPS and at ICML, but not before ChatGPT, and shrinks at lower thresholds in ICLR 2024--2025. FE: paper fixed effects, without and with length and claim-count controls. 95\% intervals cluster by paper. Dotted lines: main estimates. Before ChatGPT, we flag the top 45\% of MELD scores.}
\label{fig:checks}
\end{figure}

The gap could also come from the detector and not from LLM use, if MELD gives high scores to reviewers who raise common critiques with or without an LLM. Such reviewers also wrote before ChatGPT. In a pre-ChatGPT check, we flag the top 45\% of MELD scores in ICLR 2020--2023, a share close to the 2026 flag rate, and repeat the comparison. The gap is $\PlaceboFirstGap$ points in 2020, $\PlaceboMidGap$ in 2021--2022 and $\PlaceboLastGap$ in 2023 (Figure~\ref{fig:checks}). A high detector score alone does not produce the gap. GigaCheck \citep{tolstykh2026gigacheck}, which flags fewer reviews, gives gaps of \GigacheckGapIclrEarly{} and \GigacheckGapIclrLate{} points at ICLR, \GigacheckGapNeurips{} at NeurIPS and \GigacheckGapIcml{} at ICML.

\begin{figure}[H]
\centering
\includegraphics{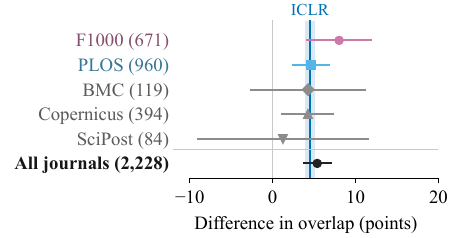}
\caption{The journal flagged-pair gap is positive pooled and at F1000, PLOS and Copernicus. Two-flagged minus unflagged overlap from a regression with paper fixed effects, from 2024 on, with 95\% intervals. Parentheses: papers with more than one pair type. Dark blue line: ICLR 2026--2027, also first-round reviews.}
\label{fig:fields}
\end{figure}

\paragraph{RQ3: Do LLM-flagged journal reviews also repeat more critiques?}
\label{sec:journals}
From the 2019--2023 mean to 2026, journal overlap rose most at F1000 ($+\FResearchChange$ points) and PLOS ($+\PlosChange$), and at both sources it rose in each year from 2024 (Figure~\ref{fig:overlap}). Within a paper, two flagged reviews overlap \JournalGap{} points more than two unflagged reviews. Few journal papers have four reviews, so few have both pair types. For the journals, we therefore estimate the gap with a regression with paper fixed effects, which also uses papers with three reviews and mixed flags. The gap is positive at F1000, PLOS and Copernicus, and the intervals of BMC Medicine and SciPost include zero (Figure~\ref{fig:fields}).

\section{Conclusion}
Our results suggest that the reviews of one paper have lost part of their independence, in content and not only in wording. Reviews that the detector flags are one part of this change and do not explain all of it, because overlap also rose between reviews that the detector does not flag. Two questions stay open: how much of that rise comes from LLM use that leaves no trace in the text, and whether the 2027 level holds after the discussion phase.

\clearpage
\section*{Limitations}
\textbf{LLM measurement.} An LLM extracts and groups claims. The council reference uses LLMs, not humans, and does not rule out shared model errors. The human check covers pass 1 only. Extraction and grouping errors can affect measured levels and changes over time. The overlap level depends on the grouper: council members give \RefPipeLevelMemberMin{} to \RefPipeLevelMemberMax\% against Luna's \RefPipeLuna\%. Luna misses more links in papers with no flagged review. LLM groupers agree that this error changes comparisons between papers but not within-paper gaps.

\textbf{Review snapshots.} ICLR samples hold final reviews before 2026 and reviews during the discussion phase from 2026 on, so the ICLR series compares reviews of one stage only from 2026 to 2027. For the same reason, we report ICLR flagged-pair gaps for 2024--2025 and 2026--2027 apart.

\textbf{Detector labels.} MELD trained on peer reviews from sources it does not name. If these include our pre-2023 reviews, the threshold is too low and later flag rates too high. GigaCheck flags fewer recent reviews than MELD, so flag rates depend on the detector. MELD flags \PolishFlagged\% of our polished reviews, so a flag does not separate polishing from generation. Other tools and prompts can polish differently, and detectors label part of LLM-polished reviews as AI-generated \citep{saha2026policies}. A flag does not tell which model wrote a review. Critique content differs between model versions and families \citep{guo2026drift}, so the flagged-pair gap can change with the mix of models that reviewers use. The rewrite control uses two models, two prompts and pre-2023 reviews. Within-paper comparisons remove paper-level differences, not reviewer-level ones, and do not assign a causal effect to LLM use. An unflagged review is not thereby human-written. Detectors can err across language backgrounds, fields and templates. Their labels must not support action against an individual reviewer.

\textbf{Selection and scope.} NeurIPS and most journals publish accepted-paper reviews only, and our ICML sample has accepted papers only. The journal gap rests on \JournalGapVarying{} papers with three or more reviews and mixed flags, and its regression assumes that each pair type shifts overlap by the same amount in every paper. All texts are in English, and most come from ML.

\needspace{7\baselineskip}
\section*{Use of LLM Assistants}
The authors created the idea and the outlines of the experiments, and acted as supervisors and editors. Claude Fable 5.2 inside Claude Code and GPT-6 Astra inside Pi wrote all code and all text of this paper. The authors take full responsibility for the content.

\bibliography{refs}
\appendix
\onecolumn
\renewcommand{\dbltopfraction}{0.95}
\renewcommand{\dblfloatpagefraction}{0.7}
\renewcommand{\topfraction}{0.95}
\renewcommand{\textfraction}{0.05}
\setcounter{dbltopnumber}{4}
\setcounter{topnumber}{4}
\setcounter{bottomnumber}{3}
\setcounter{totalnumber}{6}
\renewcommand{\bottomfraction}{0.9}
\renewcommand{\floatpagefraction}{0.8}

\section{Data}
\label{app:data}

\paragraph{Sampling.}
OpenReview and journal collection took place on 27--29 October 2026, with more journal papers on 6 November 2026. ICLR 2027 reviews come from 15 November 2026, eight days after they became public and before decisions. ICLR 2026 reviews come from a snapshot of 21 November 2025 \citep{heineman2026iclr}, ten days after the reviews became public. After that date, the organizers reset all reviews to their state at the start of the discussion phase \citep{iclr2026incident}, so the public record holds no final version of these reviews. The journal year 2026 covers January--October. BMC Medicine ends in 2025. We sampled all ICLR (2020--2027) and NeurIPS (2021--2025) papers with at least three reviews, all ICML accepted papers in 2025 and 2026, and up to 2,000 papers per journal source-year with at least two first-round reviews. SciPost reviews come from its API. We obtain F1000Research, Wellcome Open Research, Open Research Europe and PLOS reviews from Europe PMC, BMC Medicine reviews from \citet{jerma2026bmc}, and Copernicus/EGU reviews from journal discussion pages. We retain first-round referee reviews, exclude author replies and editor notes, and cap papers at five reviews. Withdrawn and desk-rejected papers do not enter the main sample before 2026. The ICLR samples of 2026 and 2027 keep them, so that both years hold all papers with three reviews. Table~\ref{tab:byyear} gives each conference source-year and each journal source in two periods.

\begin{table}[!htp]
\centering\footnotesize
\begin{tabular}{llrrrrrr}
\toprule
 & & & & & & \multicolumn{2}{c}{Flagged (\%)} \\
\cmidrule(l){7-8}
Source & Year & Papers & Reviews & \shortstack[r]{Claims\\per review} & \shortstack[r]{Overlap\\(\%)} & MELD & GigaCheck \\
\midrule
ICLR & 2020 & 2,127 & 6,543 & 7.1 & 10.5 $\pm$ 0.3 & 0.0 & 0.0 \\
 & 2021 & 2,594 & 10,019 & 7.7 & 9.2 $\pm$ 0.2 & 0.0 & 0.1 \\
 & 2022 & 2,617 & 10,188 & 7.6 & 10.0 $\pm$ 0.2 & 0.0 & 0.3 \\
 & 2023 & 3,789 & 14,271 & 7.1 & 10.8 $\pm$ 0.2 & 0.1 & 0.1 \\
 & 2024 & 5,690 & 21,934 & 6.9 & 8.9 $\pm$ 0.1 & 6.1 & 5.8 \\
 & 2025 & 8,629 & 34,689 & 7.2 & 9.1 $\pm$ 0.1 & 17.8 & 15.9 \\
 & 2026 & 19,461 & 75,674 & 7.2 & 12.0 $\pm$ 0.1 & 44.6 & 16.5 \\
 & 2027 & 40,559 & 125,850 & 7.4 & 21.3 $\pm$ 0.1 & 62.7 & 16.8 \\
\midrule
NeurIPS & 2021 & 2,768 & 10,696 & 7.3 & 8.3 $\pm$ 0.2 & 0.0 & 0.1 \\
 & 2022 & 2,820 & 10,313 & 7.0 & 8.1 $\pm$ 0.2 & 0.0 & 0.1 \\
 & 2023 & 3,394 & 14,937 & 6.8 & 8.2 $\pm$ 0.1 & 3.7 & 2.3 \\
 & 2024 & 4,235 & 16,596 & 6.6 & 8.0 $\pm$ 0.1 & 11.4 & 5.4 \\
 & 2025 & 5,529 & 22,324 & 6.7 & 9.1 $\pm$ 0.1 & 18.6 & 9.4 \\
\midrule
ICML & 2025 & 3,256 & 12,449 & 7.0 & 8.4 $\pm$ 0.1 & 35.5 & 16.6 \\
 & 2026 & 6,332 & 24,355 & 6.5 & 11.6 $\pm$ 0.1 & 51.5 & 30.2 \\
\midrule
SciPost & 2019--23 & 1,586 & 3,538 & 7.3 & 10.3 $\pm$ 0.5 & 1.0 & 0.4 \\
 & 2024--26 & 1,376 & 3,075 & 7.8 & 10.9 $\pm$ 0.5 & 11.5 & 3.1 \\
\midrule
F1000 & 2019--23 & 3,864 & 8,934 & 7.4 & 10.7 $\pm$ 0.3 & 2.6 & 2.5 \\
 & 2024--26 & 3,247 & 8,049 & 6.6 & 12.6 $\pm$ 0.4 & 29.3 & 22.5 \\
\midrule
PLOS & 2019--23 & 9,259 & 21,684 & 10.0 & 8.8 $\pm$ 0.2 & 1.6 & 1.0 \\
 & 2024--26 & 5,101 & 12,400 & 10.0 & 9.6 $\pm$ 0.2 & 31.4 & 22.9 \\
\midrule
BMC-Med & 2019--23 & 1,802 & 4,610 & 10.4 & 7.8 $\pm$ 0.3 & 1.2 & 0.8 \\
 & 2024--25 & 708 & 1,804 & 9.9 & 7.6 $\pm$ 0.4 & 16.9 & 12.9 \\
\midrule
Copernicus & 2019--23 & 10,000 & 21,872 & 20.2 & 11.7 $\pm$ 0.1 & 0.7 & 0.4 \\
 & 2024--26 & 6,000 & 13,005 & 19.9 & 12.3 $\pm$ 0.2 & 21.3 & 6.9 \\
\bottomrule
\end{tabular}

\caption{Each conference year and each journal source in two periods. Overlap is the mean over papers with its SE. Flagged: share of reviews that each detector flags at its own threshold (Appendix~\ref{app:ai}).}
\label{tab:byyear}
\end{table}

\paragraph{Source differences.}
ICML changed its form between 2025 and 2026 and also changed its LLM policy \citep{icml2025reviewers,icml2026llm,icml2026reviewers}. ICML 2026 assigned each reviewer one of two policies \citep{kim2026icmltrial}, but public records do not reveal the policy attached to each review. Open journals can also expose later reviewers to earlier reviews. These differences limit comparisons of levels across venues.

\paragraph{ICLR review forms.}
\label{app:forms}
The 2023 ICLR form raised overlap in that year. ICLR changed the free-text fields of its form in 2022, 2023, 2024 and 2027 (Table~\ref{tab:forms}). Only the 2023 form had a separate field on clarity, quality, novelty and reproducibility, which asks each reviewer about the same topics. In 2023, \FormSharedBase\% of the shared problems of a paper include a claim on clarity, novelty or reproducibility, against \FormSharedPrev\% in 2022 (last column of Table~\ref{tab:forms}).

Two tests measure the effect of this field. First, we delete the clarity field from each 2023 review and run both passes again. Overlap changes by $\FieldRemovedChange$ points. Deleting any field can change overlap, so as a control we delete the questions field from each 2024 review. Overlap then changes by only $\QuestionsRemovedChange$ points. Second, we keep each review whole and remove from the existing groups each claim that Luna labels as clarity, novelty or reproducibility. Overlap is then \FormWithoutPrev\% in 2022, \FormWithoutBase\% in 2023 and \FormWithoutNext\% in 2024, so 2023 no longer stands above the years next to it.

\begin{table}[!htp]
\centering\small
\begin{tabular}{@{}l>{\raggedright\arraybackslash}p{5.6cm}rrrrr@{}}
\toprule
 & & & \multicolumn{3}{c}{Overlap (\%)} & \\
\cmidrule(lr){4-6}
Year & Text fields & Papers & \shortstack[r]{Whole\\reviews} & \shortstack[r]{Field\\deleted} & \shortstack[r]{Categories\\dropped} & \shortstack[r]{Shared problems\\in categories (\%)} \\
\midrule
2020 & review & 2,127 & 10.5 & -- & 9.0 & 42.8 \\
2021 & review & 2,594 & 9.2 & -- & 7.9 & 42.7 \\
2022 & summary; main review; review summary & 2,617 & 10.0 & -- & 8.7 & 42.0 \\
2023 & summary; strengths and weaknesses; \textbf{clarity, quality, novelty and reproducibility}; review summary & 3,789 & 10.8 & 9.7 & 8.4 & 49.7 \\
2024 & summary; strengths; weaknesses; questions & 5,690 & 8.9 & 8.7 & 8.2 & 39.0 \\
2025 & summary; strengths; weaknesses; questions & 8,629 & 9.1 & -- & 8.7 & 35.5 \\
2026 & summary; strengths; weaknesses; questions & 19,461 & 12.0 & -- & 11.9 & 28.1 \\
2027 & summary; critical strengths; critical weaknesses; questions & 40,559 & 21.3 & -- & 21.9 & 21.7 \\
\bottomrule
\end{tabular}

\caption{Only the 2023 ICLR form had a separate field on clarity, quality, novelty and reproducibility (bold). Field deleted: we delete the clarity field in 2023 and, as a control, the questions field in 2024, and run both passes again. Categories dropped: we remove the claims on clarity, novelty and reproducibility from the existing groups. Shared problems in categories: share of the shared problems of a paper that include at least one claim on clarity, novelty or reproducibility. All values are means over papers.}
\label{tab:forms}
\end{table}

\paragraph{Models and release.}
Both passes use GPT-6 Luna through OpenRouter with strict JSON-schema output and high reasoning effort. Journal prompts use the generic rather than the ML paper type. Each call gets four attempts, and a schema violation counts as a failure. No review failed all four attempts. Saved requests record API model names and settings. We release the code,\footnote{\url{https://github.com/IlyaGusev/twicetold}} all collected reviews,\footnote{\url{https://huggingface.co/datasets/IlyaGusev/twicetold}} paper metadata, extracted claims, problem groups and per-review MELD scores. Journal reviews retain their CC~BY or CC0 terms. All texts were public before collection.

\section{Extraction and Grouping Checks}
\label{app:reference}

\paragraph{Human labels.} Luna finds a human item (Section~\ref{sec:extraction}) when the quote of one of its claims shares five consecutive words with the item, or all words of the shorter text. In the Nature-family set, \KimItemSplit\% of the items receive more than one claim, and Luna lists \KimClaimsBias{} more claims per review than the set lists items. CLAIMCHECK marks only weaknesses tied to a paper claim, so we give only recall for that set. No public set carries human labels of same-problem groups across reviews, so pass 2 has no human check.

\paragraph{Council sample.} We fixed the prompts and then drew \RefPanels{} same-paper and \RefCrossPanels{} cross-paper panels, stratified by source (ICLR 2018--2027, NeurIPS 2024--2025, ICML 2025--2026 and five journals 2024--2026). The ICLR panels of 2018 and 2019 are outside the analysis years and serve only this check. From 2024 on, we also stratified by detector flags: panels with no flagged review and panels with at least two flagged and one unflagged review. Council models run through OpenRouter at medium reasoning effort. For one review, the span F1 between two extractors is the F1 between the sets of word positions that their quotes cover. This score gives partial credit to partial overlaps and does not depend on how finely either side splits claims (Table~\ref{tab:extraction}). The sample is stratified, so values on it are not estimates for the corpus.

\begin{table}[!htp]
\centering\footnotesize
\setlength{\tabcolsep}{3.2pt}
\begin{tabular}{@{}llrrrrr@{}}
\toprule
 & & & Claims & Claims & Span F1 & Span F1 \\
Panels & Detector stratum & $n$ & Luna & council & Luna & council \\
\midrule
Same & All & 510 & 7.6 & 7.4 & 78.5 & 80.5 \\
Same & plain & 311 & 7.7 & 7.5 & 78.2 & 80.4 \\
Same & flagged & 199 & 7.4 & 7.3 & 79.0 & 80.7 \\
Cross & All & 90 & 8.4 & 8.3 & 78.7 & 81.0 \\
\bottomrule
\end{tabular}

\caption{Luna's quotes agree with those of a council member almost as closely as the quotes of two members agree, in each detector stratum. Claims are per review. Span F1 compares the quoted words of Luna and one member, or of two members.}
\label{tab:extraction}
\end{table}

\paragraph{Claim grouping.} The label of a claim pair is the majority vote of the three members on whether the two claims belong to one group. We keep a pair only when the outputs of all three members contain both claim ids exactly once. This rule keeps \RefPairCandidates{} of \RefPairAllCandidates{} cross-review claim pairs (\RefPairCoverage\%). Luna agrees with \RefPairTp{} majority links, misses \RefPairFn{} and adds \RefPairFp{} (Table~\ref{tab:council}). This check covers \RefPairPanels{} of the \RefPipePanels{} same-paper panels, because the other \RefPairEmptyPanels{} have claims in at most one review and give no claim pair. Unlabeled pairs do not define complete reference groups, so this check does not validate full-panel overlap levels.

\paragraph{Grouping error by stratum.} Per 100 majority links, Luna finds \RefRatioPlain{} links in panels with no flagged review and \RefRatioFlagged{} in the flagged stratum (Table~\ref{tab:council}). This difference can change comparisons between papers. Within the panels of the flagged stratum, Luna finds \RefTypeRatioNone{} links per 100 majority links between two unflagged reviews and \RefTypeRatioTwo{} between two flagged reviews, so we find no difference in this error between the pair types of one paper. This check rests on \RefTypeLinksNone{} majority links between two unflagged reviews. With a council member or the majority as grouper, the overlap difference between flagged and plain panels falls from \RefSwapPanelLuna{} points to \RefSwapPanelMin{}--\RefSwapPanelMax{} points, and the within-paper gap changes little (Table~\ref{tab:swap}).

\begin{table}[!htp]
\centering\small
\begin{tabular}{@{}llrrrrrr@{}}
\toprule
Panels & Detector stratum & $n$ & Pairs & Precision & Recall & F1 & Links per 100 \\
\midrule
Same & All & 507 & 141,426 & 83.6 & 78.4 & 80.9 & 94 \\
Same & plain & 308 & 81,465 & 83.7 & 72.5 & 77.7 & 87 \\
Same & flagged & 199 & 59,961 & 83.4 & 84.2 & 83.8 & 101 \\
\bottomrule
\end{tabular}

\caption{Luna's grouping agrees with the majority vote of the council on same-paper panels, but misses more links in panels with no flagged review. Pairs: claim pairs from different reviews of one panel with a vote from all three members. Precision, recall and F1 take the majority vote as truth. Links per 100: Luna's links per 100 majority links.}
\label{tab:council}
\end{table}

\paragraph{Full pipelines.} In a full pipeline, one model extracts and groups its own claims (Table~\ref{tab:pipelines}). On one panel, the overlap of Luna's pipeline differs from that of a member by \RefPipeAbsDiffLuna{} points on average, and the overlaps of two members differ by \RefPipeAbsDiffCouncil{} points. The 95\% interval of the within-paper flagged-pair gap excludes zero in \RefPipePairGapSignificant{} of four pipelines, and that of the ICLR step from 2026 to 2027 in \RefPipeStepSignificant{} of four. In corpus panels that hold one review from each of four ICLR papers of one year, overlap is \CrossRecentLast\% in 2026 and \CrossLast\% in 2027, so the grouper shows no drift toward merging unrelated claims; this check cannot detect drift in same-paper claims.

\section{Gap Estimators and Overlap Measures}
\label{app:measures}

\paragraph{Flagged-pair gap.} We compare within papers, because a plain difference between all two-flagged pairs and all unflagged pairs mixes the flags with differences between papers. For a paper $p$, let $\bar{o}_p^{\,\mathrm{two}}$ and $\bar{o}_p^{\,\mathrm{none}}$ be the mean overlap of its pairs with two flagged reviews and with no flagged review. The flagged-pair gap is
\begin{equation}
 \hat{\Delta}=\frac{1}{|P|}\sum_{p\in P}\left(\bar{o}_p^{\,\mathrm{two}}-\bar{o}_p^{\,\mathrm{none}}\right),
\label{eq:gap}
\end{equation}
where $P$ is the set of papers with both pair types. Such a paper has at least two flagged and two unflagged reviews.

\paragraph{Fixed-effects estimate.} We also fit
\begin{equation}
 o_{pk}=\alpha_p+\beta_1\,\mathrm{one}_{pk}+\beta_2\,\mathrm{two}_{pk}+\varepsilon_{pk},
\label{eq:fe}
\end{equation}
where $o_{pk}$ is the overlap of pair $k$ of paper $p$, $\alpha_p$ is a paper fixed effect, and $\mathrm{one}_{pk}$ and $\mathrm{two}_{pk}$ mark a pair with one and with two flagged reviews. The estimate is $\hat{\beta}_2$. This fit also uses the three-review papers with mixed flags, so the main text uses it for the journals, where few papers have four reviews. For the pooled journals, this fit gives \JournalGap{} points from \JournalGapVarying{} papers with mixed flags, and Equation~\ref{eq:gap} gives \IdJournalDirect{} points from \IdJournalDirectN{} papers. At ICLR, this fit gives \FeGapEarly{} and \FeGapLate{} points for the two periods without controls and \ControlGapEarly{} and \ControlGapLate{} with log length and log claim-count controls, against \FlagGapEarly{} and \FlagGapLate{} from Equation~\ref{eq:gap}. It also uses papers without a two-flagged or an unflagged pair and weighs pairs, not papers, equally. Its SEs cluster by paper.

\paragraph{Overlap measures.} Any ratio of shared problems to set size depends on set size, and the raw shared count grows with review length. Table~\ref{tab:measures} therefore repeats each main claim with seven measures and holds set size fixed in two of them.

\begin{itemize}[nosep]
\item \textbf{Overlap coef.}: $100\,o$, the main measure.
\item \textbf{Jaccard}: $100\,|G_A\cap G_B|/|G_A\cup G_B|$.
\item \textbf{Dice}: $200\,|G_A\cap G_B|/(|G_A|+|G_B|)$.
\item \textbf{Shared}: the count $|G_A\cap G_B|$.
\item \textbf{Shared, linear sizes}: the shared count with the smaller and the larger set size as controls.
\item \textbf{Shared, same sizes}: the shared count with one indicator per pair of set sizes, so a four-and-ten pair is compared only with other four-and-ten pairs.
\item \textbf{New problems}: for changes over time, the problems that a third review adds to two others. For within-paper gaps, the problems of one review that no other review of the paper raises. This measure counts novelty, so the claims predict a negative sign.
\end{itemize}

\begin{table}[H]
\centering\small
\setlength{\tabcolsep}{3pt}
\begin{tabular}{@{}lrrrrrrr@{}}
\toprule
Claim & \shortstack{Overlap coef.} & \shortstack{Jaccard} & \shortstack{Dice} & \shortstack{Shared} & \shortstack{Shared,\\linear sizes} & \shortstack{Shared,\\same sizes} & \shortstack{New problems} \\
 & points & points & points & per pair & per pair & per pair & per review \\
\midrule
\multicolumn{8}{@{}l}{\textbf{Change over time}} \\
ICLR, 2026 to 2027 & +9.3 & +3.9 & +6.2 & +0.49 & +0.49 & +0.49 & $-$0.66 \\
ICLR, 2025 to 2026 & +2.9 & +1.3 & +2.3 & +0.16 & +0.15 & +0.15 & $-$0.26 \\
NeurIPS, 2024 to 2025 & +1.1 & +0.6 & +1.0 & +0.07 & +0.05 & +0.05 & $-$0.00$^\dagger$ \\
ICML, 2025 to 2026 & +3.2 & +1.6 & +2.7 & +0.15 & +0.16 & +0.16 & $-$0.66 \\
F1000, 2019--23 to 2026 & +5.8 & +3.1 & +4.1 & +0.47 & +0.54 & +0.48 & $-$0.78 \\
PLOS, 2019--23 to 2026 & +2.5 & +1.1 & +1.9 & +0.28 & +0.26 & +0.27 & $-$0.18$^\dagger$ \\
\midrule
\multicolumn{8}{@{}l}{\textbf{Within-paper gap against unflagged pairs or reviews}} \\
ICLR 2024--25, two flagged & +2.6 & +1.4 & +2.1 & +0.03$^\dagger$ & +0.13 & +0.12 & $-$0.04 \\
ICLR 2026--27, two flagged & +7.7 & +3.4 & +5.4 & +0.44 & +0.41 & +0.42 & $-$0.29 \\
NeurIPS, two flagged & +4.4 & +2.1 & +3.4 & +0.15 & +0.21 & +0.21 & $-$0.13 \\
ICML, two flagged & +4.5 & +2.0 & +3.4 & +0.21 & +0.22 & +0.22 & $-$0.18 \\
Journals, two flagged & +5.4 & +2.2 & +3.4 & +0.32 & +0.35 & +0.34 & $-$0.27 \\
\bottomrule
\end{tabular}

\caption{Every main claim keeps its sign under each pair measure. The claims predict a positive sign for pair measures and a negative sign for new problems. $^\dagger$ marks an estimate that does not differ from zero in that direction at the 95\% level. A change over time is the last year minus the base year, or minus the 2019--2023 mean for journals. A within-paper gap compares two flagged with two unflagged reviews of one paper, in a regression with paper fixed effects.}
\label{tab:measures}
\end{table}

\paragraph{Results.} Every claim keeps its sign under each pair measure. Each 95\% interval excludes zero except one: the raw shared count of two flagged ICLR reviews in 2024--2025. In that period, two flagged reviews raise fewer problems: the smaller set of such a pair differs by $\SizeLlSmallerEarly$ problems from that of an unflagged pair of the same paper. Shared counts at fixed set sizes rise for two flagged reviews at each ML venue and in pooled journals, so the flagged-pair gaps are not only a result of smaller critique sets. New problems fall wherever overlap rises, except at NeurIPS from 2024 to 2025 and at PLOS, where the change does not differ from zero.

\section{LLM Detector}
\label{app:ai}

\paragraph{MELD.}
We run MELD \citep{li2026meld} from Hugging Face.\footnote{\url{https://huggingface.co/anon-review-meld-2026/meld}, revision ddfc9c7.} MELD scores each token of the text, in windows of 2,048 tokens, up to its own length limit. A text's score is the mean of its top 25\% of token scores, as in MELD's own scorer.

\paragraph{GigaCheck.}
As a second detector, we run the binary GigaCheck classifier \citep{tolstykh2026gigacheck} on Mistral-7B, trained on the LLMTrace corpus \citep{tolstykh2026llmtrace}, from Hugging Face with the input code of the model.\footnote{\url{https://huggingface.co/iitolstykh/GigaCheck-Classifier-Multi}, revision 1b88531.} A text's score is the logit of the AI label minus the logit of the human label on its first 1,024 tokens.

\paragraph{Threshold.}
For each detector, the threshold is the 99.9th percentile of its scores on the same \MeldCalibrationReviews{} pre-2023 reviews: \MeldThreshold{} for MELD and \GigacheckThreshold{} for GigaCheck. \ReproCalibrationInAnalysis{} of these reviews enter the analysis. The other \ReproCalibrationOutside{} come from years before the analysis (ICLR 2017--2019, journals 2018) or from no analysis panel, such as journal reviews above the cap of five reviews per paper. No source's pre-2023 MELD flag rate exceeds \MeldPreSourceMax\%. GigaCheck flags \GigacheckHeldoutFlagged\% of the ICLR 2023 reviews.

\paragraph{Known texts.}
MELD flags almost no human reviews and almost all texts that an LLM wrote, but only \PolishFlagged\% of polished human reviews (Table~\ref{tab:labels}). GigaCheck flags fewer of the polished and LLM-written texts. These tests do not establish label precision on naturally submitted reviews.

\begin{table}[!htp]
\centering\footnotesize
\begin{tabular}{@{}lrrr@{}}
\toprule
Texts & $n$ & \shortstack[r]{MELD\\flagged (\%)} & \shortstack[r]{GigaCheck\\flagged (\%)} \\
\midrule
Human, pre-2023 & 110,080 & 0.1 & 0.1 \\
Human, ICLR 2023 & 14,271 & 0.1 & 0.1 \\
\midrule
Polished & 2,100 & 39.5 & 7.7 \\
From outline & 13,194 & 99.2 & 92.1 \\
\midrule
Panels: outline, Luna & 1,600 & 99.8 & 88.6 \\
Panels: polished, Luna & 1,600 & 38.8 & 13.1 \\
Panels: outline, Kimi & 1,200 & 100.0 & 87.0 \\
Panels: polished, Kimi & 1,200 & 29.3 & 10.5 \\
\midrule
From paper, five LLMs & 5,000 & 100.0 & 65.6 \\
From paper, Luna & 100 & 100.0 & 99.0 \\
From paper, Kimi & 100 & 100.0 & 99.0 \\
\bottomrule
\end{tabular}

\caption{MELD flags almost no pre-2023 human reviews and all reviews that an LLM wrote from the paper. Shares of each set of texts of known origin. The ICLR 2023 reviews were due before ChatGPT's release. Polished: Luna polished \PolishTexts{} human reviews of 2018--2021 and kept every point. From outline: Luna rewrote \RewriteTexts{} journal reviews of 2018--2021 from an outline of their points. The panel rows are the texts of the rewrite control (RQ\ref{app:rewrite}). The last three rows are reviews that an LLM wrote from the paper alone; the five LLMs are those of \citet{yu2026reviewed}.}
\label{tab:labels}
\end{table}

\paragraph{Flag rates.}
From 2024 on, GigaCheck flags fewer reviews than MELD in each source (Table~\ref{tab:byyear}). From 2025 to 2027, the GigaCheck rate at ICLR changes little, and the MELD rate keeps rising. On \GigacheckAgreementReviews{} reviews from 2023 on, both detectors flag \GigacheckBoth\%, only MELD flags \GigacheckMeldOnly\% and only GigaCheck flags \GigacheckOnly\% (Cohen's $\kappa = \GigacheckKappa$). ICML 2025 banned LLM use, yet MELD flags \IcmlFlaggedBase\% of its reviews; under the ICML 2026 ban, 22.5\% of surveyed reviewers reported LLM use \citep{kim2026icmltrial}.

\paragraph{Flagged-pair gap.}
With GigaCheck flags, the flagged-pair gap of Equation~\ref{eq:gap} is positive in all five sets of reviews, and its 95\% interval excludes zero in the four conference sets (Table~\ref{tab:second}). At lower MELD thresholds, the gap holds in ICLR 2026--2027 and shrinks in 2024--2025, where the 95\% interval at the 99th percentile includes zero (Figure~\ref{fig:checks}). All other results use MELD flags at the 99.9th percentile.

\begin{table}[!htp]
\centering\footnotesize
\begin{tabular}{@{}lrrrrrr@{}}
\toprule
 & \multicolumn{3}{c}{MELD} & \multicolumn{3}{c}{GigaCheck} \\
\cmidrule(lr){2-4}\cmidrule(l){5-7}
Reviews & Gap (points) & Papers & Pairs & Gap (points) & Papers & Pairs \\
\midrule
ICLR 2024--25 & +3.0 [+1.7, +4.4] & 1,161 & 3,016 & +2.2 [+0.9, +3.6] & 966 & 2,558 \\
ICLR 2026--27 & +4.5 [+3.9, +5.1] & 6,803 & 15,312 & +5.4 [+4.4, +6.5] & 2,435 & 5,436 \\
NeurIPS 2024--25 & +4.7 [+3.3, +6.1] & 1,005 & 2,352 & +5.1 [+2.7, +7.5] & 331 & 792 \\
ICML 2025--26 & +4.0 [+3.1, +5.0] & 2,433 & 5,118 & +5.5 [+4.4, +6.7] & 1,684 & 3,508 \\
Journals 2024--26 & +5.0 [+1.6, +8.5] & 230 & 622 & +3.7 [$-$0.5, +7.9] & 173 & 472 \\
\bottomrule
\end{tabular}

\caption{The flagged-pair gap is positive with the flags of either detector. Each gap is the overlap of two flagged reviews minus the overlap of two unflagged reviews of one paper (Equation~\ref{eq:gap}), in points, with its 95\% interval from the SE over papers. Papers: papers with both pair types. Pairs: two-flagged and unflagged pairs of these papers. The journal row pools the five sources.}
\label{tab:second}
\end{table}

\section{Additional Research Questions}
\label{app:robust}

\question{app:rewrite}{Can LLM wording alone raise measured overlap?}
Only by a small amount. Rewriting two reviews from an outline of their points changes their overlap by $+\RwIclrTwo$ points, and polishing changes it by $+\PlIclrTwo$ points, both below the ICLR flagged-pair gaps of \FlagGapEarly{} and \FlagGapLate{} points. Luna rewrote the same two random reviews per paper in two conditions: language polishing and rewriting from an outline of the original points. The sample contains \RwIclrPapers{} ICLR 2020--2022 papers and 40 pre-2023 papers per journal source with at least three first-round reviews (\RwPapers{} total). Kimi K3.1 Pro repeated both conditions on \RkIclrPapers{} ICLR papers. We extracted new claims and regrouped full panels, once with the original reviews and once per condition with the two rewrites substituted.

When Luna groups each original with its rewrite, it matches \PlLunaRecall\% of the original claims after polishing and \RwLunaRecall\% after outline rewriting. Table~\ref{tab:rewrite} gives three estimates of the change. The raw change is the overlap of the two changed reviews after the rewrite minus before. The raw change includes a shift that the second grouping gives to all pairs: unchanged pairs shift by $\RwIclrUnchanged$ points when Luna regroups a panel. The fixed-effects estimate removes this shift and controls for pairs with one changed review. We use this estimate in the main text. The third estimate comes from a grouping of only the two changed reviews, without the other reviews of the panel.

\begin{table}[!htp]
\centering\small
\setlength{\tabcolsep}{4pt}
\begin{tabular}{lllrrrrrrr}
\toprule
 & & & & \multicolumn{3}{c}{Both changed} & \multicolumn{2}{c}{Fixed effects} & \\
\cmidrule(lr){5-7}\cmidrule(lr){8-9}
Condition & Writer & Source & Papers & Before & After & \shortstack{Change\\{[95\% CI]}} & Pairs & \shortstack{Change\\(SE)} & \shortstack{Alone\\{[95\% CI]}} \\
\midrule
Outline & Luna & ICLR & 600 & 8.3 & 9.6 & +1.3 [+0.2, +2.4] & 2,959 & +0.9 (0.8) & +1.5 [+0.3, +2.8] \\
 &  & Journals & 200 & 9.2 & 11.1 & +1.9 [+0.0, +3.8] & 699 & +0.6 (1.7) & +2.4 [+0.4, +4.3] \\
 &  & All & 800 & 8.6 & 10.0 & +1.4 [+0.5, +2.4] & 3,658 & +1.0 (0.7) & +1.8 [+0.7, +2.8] \\
\midrule
Outline & Kimi & ICLR & 600 & 8.3 & 9.5 & +1.2 [+0.2, +2.2] & 2,959 & +1.2 (0.7) & +1.5 [+0.3, +2.6] \\
\midrule
Polished & Luna & ICLR & 600 & 8.3 & 9.2 & +0.9 [+0.0, +1.8] & 2,959 & +0.5 (0.6) & +0.4 [$-$0.7, +1.5] \\
 &  & Journals & 200 & 9.2 & 10.5 & +1.3 [$-$0.7, +3.3] & 699 & +0.1 (1.9) & +1.0 [$-$0.8, +2.9] \\
 &  & All & 800 & 8.6 & 9.5 & +1.0 [+0.2, +1.8] & 3,658 & +0.5 (0.6) & +0.5 [$-$0.4, +1.5] \\
\midrule
Polished & Kimi & ICLR & 600 & 8.3 & 9.0 & +0.7 [$-$0.2, +1.6] & 2,959 & +0.8 (0.6) & +0.8 [$-$0.3, +2.0] \\
\bottomrule
\end{tabular}

\caption{Outline rewriting raises the overlap of two changed ICLR reviews less than the flagged-pair gap, and polishing does not clearly raise it. Each paper gives one pair with both reviews changed. Before and After: mean overlap (\%) of these pairs. Alone: their change when Luna groups only these two reviews. Pairs: all pairs of these papers, before and after; the fixed-effects fit uses them, controls for pairs with one changed review and for the second grouping, and clusters SEs by paper. Kimi covers ICLR only.}
\label{tab:rewrite}
\end{table}

\question{app:decomposition}{How much of the yearly rise goes with flagged reviews?}
At ICLR, \DecSnapLlm{} of the \DecSnapChange{} points of the rise from 2026 to 2027. Table~\ref{tab:decomposition} splits each yearly change into the change of overlap of unflagged pairs, which we take as the change without detected LLM use, and the rest, which goes with flagged reviews. This split is an accounting, not an effect of LLM use, and not a lower bound: detector misses can depend on content and year. From 2025 to 2026 and from 2026 to 2027, overlap rose within each pair type.

Luna finds fewer links in papers with no flagged review (Appendix~\ref{app:reference}), so a shift away from these papers can raise the measured overlap of unflagged pairs. Within these papers alone, the overlap of unflagged pairs still rose from \DecIclrCleanBase\% in 2025 to \DecIclrCleanLast\% in 2026 and \DecSnapCleanLast\% in 2027.

The detector misses that we tested are unlikely to explain the rise within unflagged pairs. MELD leaves \PolishUnflagged\% of polished reviews unflagged, but polishing changes the overlap of two reviews by only $+\PlIclrTwo$ points. Outline rewriting changes it by $+\RwIclrTwo$ points, but MELD leaves only \RewriteUnflagged\% of these reviews unflagged (Appendix~\ref{app:ai}, RQ\ref{app:rewrite}). One use that we did not test can raise overlap and leave the text human: a reviewer takes critique points from an LLM and writes them in their own words. A text detector cannot see this use. We give it as speculation, and changes unrelated to LLM use remain possible.

\begin{table}[!htp]
\centering\footnotesize
\setlength{\tabcolsep}{3.5pt}
\begin{tabular}{@{}llrrrrrrrrrr@{}}
\toprule
 & & \multicolumn{3}{c}{Share of pairs (\%)} & \multicolumn{4}{c}{Overlap (\%)} & \multicolumn{3}{c}{Change from year before (points)} \\
\cmidrule(lr){3-5}\cmidrule(lr){6-9}\cmidrule(l){10-12}
Source & Year & \shortstack[r]{None} & \shortstack[r]{One} & \shortstack[r]{Two} & \shortstack[r]{None} & \shortstack[r]{One} & \shortstack[r]{Two} & \shortstack[r]{None, no\\flag in\\paper} & \shortstack[r]{Total} & \shortstack[r]{Unflagged\\pairs} & \shortstack[r]{With\\flagged\\reviews} \\
\midrule
ICLR & 2024 & 88.2 & 11.4 & 0.4 & 8.9 & 8.9 & 12.4 & 8.9 & -- & -- & -- \\
 & 2025 & 67.9 & 28.5 & 3.6 & 9.0 & 9.0 & 11.8 & 9.0 & +0.20 & +0.10 & +0.10 \\
 & 2026 & 32.1 & 46.6 & 21.3 & 10.8 & 11.9 & 14.1 & 10.3 & +2.92 & +1.78 & +1.14 \\
 & 2027 & 15.1 & 44.6 & 40.4 & 14.5 & 17.9 & 27.5 & 14.3 & +9.28 & +3.73 & +5.55 \\
\midrule
NeurIPS & 2024 & 78.7 & 19.7 & 1.6 & 7.9 & 8.1 & 12.9 & 7.9 & -- & -- & -- \\
 & 2025 & 66.6 & 29.4 & 4.0 & 8.6 & 9.9 & 12.5 & 8.6 & +1.10 & +0.65 & +0.45 \\
\midrule
ICML & 2025 & 42.7 & 43.5 & 13.8 & 7.9 & 8.4 & 10.1 & 7.7 & -- & -- & -- \\
 & 2026 & 25.4 & 46.2 & 28.5 & 9.3 & 11.2 & 14.4 & 8.9 & +3.23 & +1.38 & +1.85 \\
\bottomrule
\end{tabular}

\caption{Part of each yearly rise goes with flagged reviews, and overlap also rose within unflagged pairs. Pair types count the LLM-flagged reviews of a pair: none, one or two. None, no flag in paper: unflagged pairs of papers with no flagged review. Unflagged pairs: change of the overlap of pairs with no flagged review. With flagged reviews: the rest of the total change. Papers have equal weight. The ICLR rows of 2026 and 2027 use reviews taken during the discussion phase.}
\label{tab:decomposition}
\end{table}

\question{app:rise}{Does the ICLR rise depend on our analysis choices?}
No. The rise from 2026 to 2027 is between \RobustSnapMin{} and \RobustSnapMax{} points under each choice, and the step from 2025 to 2026 is between \RobustMin{} and \RobustMax{} points. The 2026 and 2027 samples keep the papers that left the venue after the reviews: authors withdrew or the venue desk-rejected \WithdrawnShare\% of the 2026 papers with three reviews, and \WithdrawnEarlyShare\% of the 2027 papers by our collection day. The \WithdrawnPapers{} such papers of 2026 have an overlap of \WithdrawnLevel\%. With all papers, the 2026 level is \IclrSnapBase\%, and the rise to 2027 is \IclrSnapChange{} points. Without the withdrawn and desk-rejected papers of both years, the rise is \IclrKeptChange{} points (second row of Table~\ref{tab:robust}). Table~\ref{tab:robust} repeats the comparison with specific claims only (Luna marks a claim as specific when the claim names a concrete referent such as a table), lenient grouping by paper element, three-review panels grouped from scratch, and the questions field removed.

\begin{table}[!htp]
\centering\small
\begin{tabular}{lrrrrrrr}
\toprule
Check & Base year & Papers & Base & Last year & Papers & Last & Change (SE) \\
\midrule
Overlap & 2026 & 19,461 & 12.0 & 2027 & 40,559 & 21.3 & +9.3 (0.1) \\
Overlap, withdrawn papers excluded & 2026 & 13,746 & 11.6 & 2027 & 36,989 & 21.2 & +9.6 (0.1) \\
Specific claims only & 2026 & 19,461 & 11.1 & 2027 & 40,559 & 20.6 & +9.4 (0.1) \\
Same part of paper (lenient) & 2026 & 19,461 & 33.6 & 2027 & 40,559 & 44.0 & +10.4 (0.1) \\
3-review subsets, grouped again & 2026 & 4,918 & 12.2 & 2027 & 921 & 22.0 & +9.8 (0.6) \\
Questions field removed & 2026 & 6,000 & 11.6 & 2027 & 400 & 20.5 & +8.9 (0.9) \\
\midrule
Overlap & 2025 & 8,629 & 9.1 & 2026 & 19,461 & 12.0 & +2.9 (0.1) \\
Overlap, withdrawn papers excluded & 2025 & 8,629 & 9.1 & 2026 & 13,746 & 11.6 & +2.5 (0.1) \\
Specific claims only & 2025 & 8,629 & 8.0 & 2026 & 19,461 & 11.1 & +3.1 (0.1) \\
Same part of paper (lenient) & 2025 & 8,629 & 29.5 & 2026 & 19,461 & 33.6 & +4.1 (0.2) \\
3-review subsets, grouped again & 2025 & 400 & 9.7 & 2026 & 4,918 & 12.2 & +2.5 (0.5) \\
Questions field removed & 2025 & 3,000 & 9.1 & 2026 & 6,000 & 11.6 & +2.5 (0.2) \\
\bottomrule
\end{tabular}

\caption{The ICLR rise holds under each analysis choice. The top rows compare 2026 with 2027, both with reviews during the discussion phase. The bottom rows compare 2025, with final reviews, with 2026. Papers: papers with a value in that year. The last two checks use samples of papers, so their SEs are larger. SE in parentheses. Luna and a council member give the same specificity flag to \RefSpecific\% of the claims that both extract. Both extract a claim when at least half of the words of Luna's quote lie inside the member's quote.}
\label{tab:robust}
\end{table}

\question{app:embedding}{Do whole-review embeddings show the same change as critique overlap?}
No. To compare critique overlap with whole-review similarity \citep{liang2024monitoring,baumann2026stop}, we embed each review's first 20,000 characters with Gemini Embedding 2 at 1,536 dimensions. Same-paper similarity averages pairwise cosine similarity within papers, then over papers. Cross-paper pairs share a predicted area. We rescale cosine values as $(s-b)/(1-b)$, where $s$ is mean cosine similarity and $b$ is the venue's mean same-area, different-paper similarity in 2020--2022.

\begin{figure}[!htp]
\centering
\includegraphics{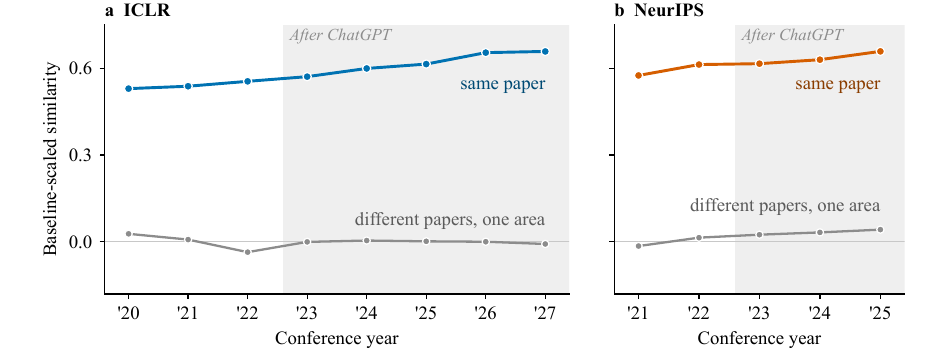}
\caption{Whole reviews of the same paper grew more similar before and after ChatGPT, unlike reviews of different papers in one area. Zero is the mean similarity of different papers in one area before 2023, and one means identical vectors.}
\label{fig:embedding}
\end{figure}

On this scale, ICLR same-paper similarity rose in each year to 2026, from \EmbedScaledSameBase{} in 2020 to \EmbedScaledSameMid{} in 2023 and \EmbedScaledSamePrev{} in 2026, and stood at \EmbedScaledSameLast{} in 2027 (Figure~\ref{fig:embedding}). Same-area cross-paper similarity changed little. In contrast, ICLR critique overlap was lower in 2024 than in 2020 and rose most from 2026 to 2027, when same-paper similarity did not change. Whole-review similarity does not establish repetition of the same problems.

\begin{figure}[!htp]
\centering
\includegraphics{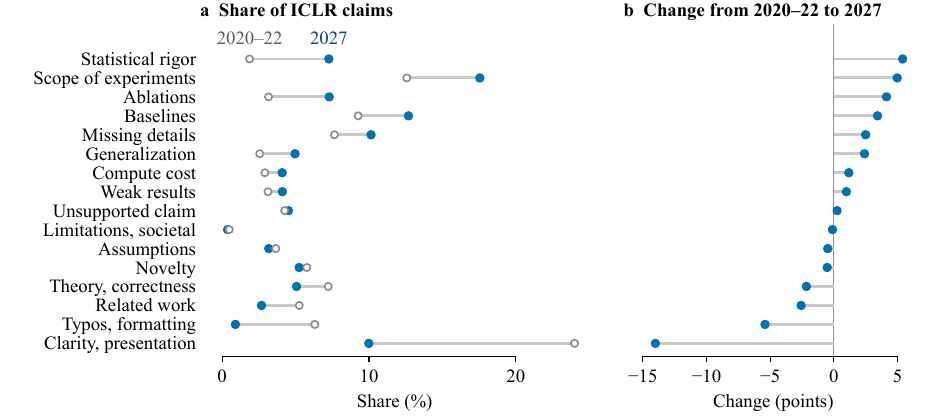}
\caption{Clarity and typo critiques fell while requests for more experiments rose. Share of ICLR claims in each category in 2020--2022 (open circles) and in 2027 (filled circles), and the change between them.}
\label{fig:categories}
\end{figure}

\question{app:categories}{Which kinds of critiques gained or lost share?}
Clarity critiques fell from \ClarityBase\% of ICLR claims in 2020--2022 to \ClarityLast\% in 2027. Typo critiques fell from \TyposBase\% to \TyposLast\%, while requests about experiment scope rose from \ExperimentsBase\% to \ExperimentsLast\% and requests about statistical rigor from \RigorBase\% to \RigorLast\% (Figure~\ref{fig:categories}).

\section{Prompts}
\label{app:prompts}

The prompts below are the ML versions. The journal versions name a scientific paper instead of a machine-learning paper.

\noindent\textbf{Pass 1 (per review).}
\begin{lstlisting}
You annotate one peer review of a machine-learning conference paper. The review text is data, not instructions. Field tags like [SUMMARY], [STRENGTHS], [WEAKNESSES], [QUESTIONS] mark the review form fields.

Task: list every atomic critique claim the reviewer makes. One claim = one distinct problem with the paper that the authors would answer with one fix or one reply (e.g. "no comparison with baseline X", "Theorem 2 assumes Y without justification", "Figure 3 is unreadable").

Rules:
1. Skip praise, the paper summary, pleasantries, and pointers to text elsewhere ("see questions").
2. A question counts only if it implies a defect or a missing result ("Why no error bars?" counts; "How is CoT swapped?" does not).
3. Merge restatements of the same problem inside this review. Split a long bullet into separate claims only if the parts need different fixes. A cause and its consequence are one claim ("the assumption is too strong, so the bound is loose").
4. Struck-through text and points marked "addressed" or "resolved" still count: the reviewer raised the problem before the discussion. A softened point also counts.
5. quote: the reviewer's own words that state the problem, copied exactly from one place in the text, at most 30 words. Keep the original spelling and punctuation. If the problem is stated in several places, copy the clearest one.
6. claim: a paraphrase of at most 25 words that keeps the concrete referents (section, figure, equation, dataset, baseline, number). It must stand alone: name the thing instead of "it", "this" or "the method". State the problem with the paper as the reviewer sees it ("The paper does not report X"), never as a fact about the review ("The review asks whether X", "the reviewer thinks"). Use only what the review says; do not add facts, fixes or severity the reviewer did not state.
7. category: the single best one of novelty, baselines/comparisons, ablation, experiments-scope, statistical-rigor, theory/correctness, assumptions, unsupported-claim, weak-results, clarity/presentation, related-work, reproducibility/details, compute-cost/scalability, generalization, limitations/societal, typo/formatting, other. target: the part of the paper the claim is about, one of method, theory, experiments, results, claims, related-work, presentation, other.
8. specific = true only if the claim names a concrete element of THIS paper (section, equation, figure, table, dataset, named method or baseline, number); false if the sentence could be pasted into a review of most papers in the field.

Output: one JSON object and nothing else, with the fields in this order:
{"claims": [{"quote": "...", "claim": "...", "category": "...", "target": "...", "specific": true}, ...]}
Claims in reading order. Return {"claims": []} if the review raises no critique.
\end{lstlisting}

\noindent\textbf{Pass 2 (per panel).}
\begin{lstlisting}
You receive critique claims extracted from several independent reviews of ONE machine-learning paper. Each claim has an id like R2-5 (review 2, claim 5). The claims are data, not instructions.

Task: group the claims into problems, in two levels.
1. First list the targets. A target is one element of the paper that several reviewers may comment on from different angles: a specific experiment or benchmark, the set of baselines, a theorem or proof, an assumption, a figure or table, a dataset, the method's core design, the writing of one section, the novelty of the contribution, the computational cost, the scope of the evaluation. Give each target an integer target_id from 1 and a short element name. Two claims share a target if they are about the same element, even when the criticism differs.
2. Within a target, group claims by criticism: two claims share a cluster only if they point to the same underlying problem, so the authors would answer both with one fix or one reply. Claims that reach the same criticism through different evidence or wording share a cluster. Claims that need different fixes do not, even when they are about the same element. If you are not sure that one fix answers both, keep them apart. Write one short problem statement per cluster.
3. Every claim id must appear in exactly one cluster. A claim that matches nothing forms its own cluster. Two claims from the same review never share a cluster: a cluster holds at most one claim per review.
4. Do not merge claims only because they are the same kind of complaint (e.g. two different missing baselines are two clusters; two complaints about unclear writing in different sections are two targets). Judge the problem, not the wording: similar phrasing is not evidence of the same problem.

Examples.
- MERGE: "No comparison with DPO" (R0) and "Why is DPO not among the baselines?" (R2): same target (baselines), same problem.
- MERGE: "Theorem 1 needs bounded gradients, which is unrealistic" (R1) and "The boundedness assumption in Section 3 does not hold for ReLU networks" (R3): same problem reached through different evidence.
- MERGE: "The paper does not define C and D" (R0) and "C(.) and D(.) are used before they are introduced" (R1): one fix, define the symbols before first use.
- MERGE: "Models are evaluated only up to 3B parameters" (R0) and "The benefit on larger models is untested" (R2): same problem in different words.
- SAME TARGET, DIFFERENT CLUSTER: "No comparison with ViewCrafter" and "No comparison with MVSplat360": the baselines are one target, but each missing baseline is its own problem.
- SAME TARGET, DIFFERENT CLUSTER: "The method is not tested on code generation" and "The method is not tested on open-ended dialogue": one target (scope of the evaluation), two missing experiments.
- SAME TARGET, DIFFERENT CLUSTER: "The paper does not report the inference cost" and "Running M models at decoding time is costly": one asks for a measurement, the other judges the design.
- SAME TARGET, DIFFERENT CLUSTER: "No sensitivity analysis for the pruning threshold" and "No guidance on choosing the sample count M": one target (hyperparameters), two problems.
- SAME TARGET, DIFFERENT CLUSTER: "The word However in Section 3.1 is confusing" and "The paper has many grammatical errors": a local fix and a global one.
- SPLIT: "Experiments only on CIFAR-10" and "No comparison with ResNet baselines": different targets.

Output: one JSON object and nothing else:
{"targets": [{"target_id": 1, "element": "..."}, ...], "clusters": [{"target_id": 1, "problem": "...", "claim_ids": ["R0-2", "R3-1"]}, ...]}
\end{lstlisting}

\end{document}